\documentclass{article}
\pdfoutput=1
\usepackage{spconf, amsmath, amsfonts, graphicx}
\usepackage{booktabs} 
\usepackage{bm,  algorithm, algorithmic}

\usepackage{dblfloatfix}  

\newtheorem{theorem}{Theorem}

\newtheorem{lemma}{Lemma}

\title{Feature Selection for multi-labeled variables via Dependency Maximization}
\name{Salimeh Yasaei Sekeh and Alfred O. Hero III
\thanks{This work was partially supported by ARO under grant W911NF-15-1-0479.}}%
\address{Department of Electrical Engineering and Computer Science\\
University of Michigan\\
1301 Beal Ave, Ann Arbor, MI, 48109, USA}
\begin{document}
\ninept
\maketitle
\begin{abstract}
Feature selection and reducing the dimensionality of data is an essential step in data analysis. In this work we propose a new criterion for feature selection that is formulated as conditional information between features given the labeled variable. Instead of using the standard mutual information measure based on Kullback-Leibler divergence, we use our proposed criterion to filter out redundant features for the purpose of multiclass classification. This approach results in an efficient and fast non-parametric implementation of feature selection as it can be directly estimated using a geometric measure of dependency, the global Friedman-Rafsky (FR) multivariate run test statistic constructed by a global minimal spanning tree (MST).  We demonstrate the advantages of our proposed feature selection approach through simulation. In addition the proposed feature selection method is applied to the MNIST data set.
\end{abstract}

\def\BX{\mathbf{X}} \def\bx{\mathbf{x}} \def\by{\mathbf{y}}
\def\BS{\mathbf{S}} \def\BZ{\mathbf{Z}} \def\BY{\mathbf{Y}}
\def\BK{\mathbf{K}}
\def\tL{\mathbf{L}}
\def\BB{\mathbf{B}}
\def\vphi{{\varphi}}
\def\rw{{\rm w}}
\def\bZ{\mathbf Z}
\def\wtf{{\widetilde f}} \def\wtg{{\widetilde g}} \def\wtG{{\widetilde G}}
\def\vphi{\varphi}
\def\rT{{\rm T}}
\def\tA{{\tt A}} \def\tB{{\tt B}} \def\tC{{\tt C}} \def\tI{{\tt I}} \def\tJ{{\tt J}} \def\tK{{\tt K}}
\def\tL{{\tt L}} \def\tP{{\tt P}} \def\tQ{{\tt Q}} \def\tS{{\tt S}}
\def\beac{\begin{array}{c}} \def\beal{\begin{array}{l}} \def\beacl{\begin{array}{cl}} \def\ena{\end{array}}
\def\bbV{\mathbb{V}}
\def\bbS{\mathbb{S}}
\def\diy{\displaystyle}
\def\bx{\mathbf{x}}
\def\rd{\rm {d}}
\def\ep{\epsilon}
\def\bbR{\mathcal{R}}
\def\bbE{\mathcal{E}}
\section{Introduction}

Feature selection has been widely investigated in various fields such as machine learning, signal processing, pattern recognition, and data science. The goal of feature selection is to select the smallest possible feature subset that preserves the information in the original high dimensional feature set.
In the classification context, the problem is to find the feature subset of minimum cardinality that preserves the information contained in the whole set of features with respect to class/label set $C=\{c_1,c_2,\ldots,c_m\}$. This problem is often solved by using a criterion that distinguishes between the relevant features and the irrelevant ones. 
There are three main types of feature selection methods: 1) wrapper \cite{KohaviandJohn1997}, 2) embedded \cite{Laletal2006}, and 3) filter methods \cite{GuyonandElisseeff2003}. Filter methods are relatively robust against overfitting, but may fail to select the best feature subset for classification or regression.
Feature selection performance is usually measured in terms of the classification error rate obtained on a testing set.
For instance consider the feature set $\BX=\{\BX^{(1)},\BX^{(2)},\ldots,\BX^{(d)}\}$ with class set $C=\{c_1,c_2,\ldots,c_m\}$. Hellman and Raviv (1970) \cite{HellmanRavin1970} obtained an upper bound, $\diy\frac{1}{2}H(C|\BX)=\diy\frac{1}{2}\left(H(C)-I(C;\BX)\right)$, on minimum Bayes classification error, where $H$ and $I$ are the entropy and the mutual information respectively \cite{Battiti1994, Vergara2014}. 
For fixed entropy $H(C)$ this upper bound is minimized when the mutual information between $C$ and $\BX$ is maximized. This intuition inspired researchers to employ $I(C;\BX)$ maximization to select the most informative feature subsets.

Intuitively, a given feature is relevant when either individually or together with other variables, it provides high level of information about the class $C$.
We measure the relevancy between features given the class label $C=c_k$ by a conditional dependency measure $R$ between $\BX^{(i)}$ and $\BX^{(j)}$, for all $k=1,2,\ldots,m$ and $1\leq i<j\leq d$.  
The proposed measure of conditional dependency between features is novel. We incorporate prior class probabilities by taking weighted average of conditional dependencies where the priors are the weights: 
\vspace{-8pt}
\begin{equation}\label{eq(1)}
\diy\sum\limits_{k=1}^m P(C=c_k)R(\BX^{(i)};\BX^{(j)}|C=c_k).\end{equation}
Here $R$ is a dependency measure and represents the relevance between two features $\BX^{(i)}$ and $\BX^{(j)}$. The objective is to find the features that have maximum conditional dependency i.e. maximizing (\ref{eq(1)}), so that they can be dropped. In other words, the dependency measure given in (\ref{eq(1)}) is sorted and features with higher total pairwise measures are filtered out. Later in this work we apply a geometric interpretation of a nonparametric measure called the $R$ metric, and show that (\ref{eq(1)}) is the conditional geometric mutual information (GMI) $I(\BX^{(i)};\BX^{(j)}|C)$, proposed in \cite{SalimehEntropy2018}. The main motivation for using GMI is the ability to estimate it directly without density estimation. The GMI estimator we propose in this paper is simple to implement and faster than plug-in approaches \cite{KSG, PPS, MNYSH2017, MSH2017} and standard pairwise class MI approaches \cite{SalimehetalAlertton2018}. The empirical estimator involves the construction of a global MST spanning over both the original data and a randomly permuted version of this data within each class.

The rest of the paper is organized as follows. Section~\ref{RW} briefly reviews related work on feature selection and dimensionality reduction. Section~\ref{GNM} defines the geometric nonparametric measure named conditional geometric mutual information (GMI). 
A novel global class-based Friedman-Rafsky statistic is proposed in Section~\ref{Estimate} and we prove that this statistic estimates the conditional GMI when the samples sizes of all classes increases simultaneously in a specific regime. Section~\ref{Experiments} is dedicated to the numerical studies and real data set experiments. Finally, Section~\ref{conclude} concludes the paper.

\section{Related Works}\label{RW}
Several methods for dimensionality reduction have been proposed in the literature. One well known linear transform for dimensionality reduction is principal component analysis (PCA) \cite{Devijver1982}. 
This technique has been studied in wide range of papers and is well-developed in various areas of Machine Learning and Signal Processing. 
Techniques like PCA are unsupervised learning methods and do not use the class labels. This is a drawback because the information from class labels is ignored. Furthermore, PCA is only sensitive to linear dependencies.
The mutual information between the class labels and the transformed data is an alternative technique that leverages the class labels and overcomes the PCA limitations \cite{Vergara2014}. 
However, plug-in based Mutual Information methods require probability density estimation \cite{Principeetal2000, MSH2017,YangMoody2000} which is computationally expensive. Berisha et al. \cite{Berishaetal2016} proposed a direct estimation of the Bayes error rate bounds. This estimation method was used to select features. We compare our method to a multi-class extension of Barisha et al.

Recently, there have been a number of attempts to non-parametrically approximate divergence measures and in particular, the mutual information, using graph-based algorithms such as minimal spanning tree (MST), \cite{Yukich, AldousSteel1992} and $k$-nearest neighbors graphs ($k$-NNG), \cite{Beardwoodetal1959, NoshadHero2018}. Lately the direct estimator based on Friedman-Rafsky (FR) multivariate test statistic \cite{FR} has received attention. This approach is constructed from the MST on the concatenated data set drawn from sufficiently smooth probability densities. Henze and Penrose \cite{HP,BerishaHero2015,Berishaetal2016} showed that the FR test is consistent against all alternatives. A further development on the graph-based approaches for multi-class classification (multi-labeled algorithms) is to construct a global graph over the entire data instead of pairwise subsets. This approach and its advantages in multi-classification problem have been studied in \cite{SalimehetalAlertton2018}. 
We extend \cite{SalimehetalAlertton2018} to estimate conditional mutual information given class labels, which is then used for feature selection.

\section{Geometric Nonparametric measure}\label{GNM}
In this section we propose a novel dependency measure called \emph{geometric conditional mutual} information.

Let $\BX^{(i)}$ be the $i$'th component of a $d$ dimensional random vector $\BX=\{\BX^{(1)},\BX^{(2)},\ldots,\BX^{(d)}\}$. Consider label variable $Y$ corresponding to class $c_1,c_2,\ldots,c_m$ that takes values $\{1,2\ldots,m\}$. Let $p_y=P(Y=y)$ for $y=1,2,\ldots,m$ such that $\diy\sum\limits_{y}p_y=1$.\\
We use notation $\pi_{ij}$  for the joint distribution $i$th and $j$th components when $\BX^{(i)}$, $\BX^{(j)}$, and $Y$  form a Markov chain, $\BX^{(i)}\rightarrow Y\rightarrow \BX^{(j)}$, i.e. random vectors $\BX^{(i)}$ and $\BX^{(j)}$ are conditionally independent given label variable $Y$,
\begin{equation}\label{markov.joint}
\pi_{ij}:=\pi(\BX^{(i)},\BX^{(j)})=\diy\sum\limits_{y} p_y f(\bx^{(i)}|y) f(\bx^{(j)}|y).
\end{equation}
The joint distribution $\BX^{(i)}$ and $\BX^{(j)}$ is given by
\begin{equation}\label{standard.joint}
f_{ij}:=f(\bx^{(i)},\bx^{(j)})=\diy\sum\limits_y p_y f(\bx^{(i)},\bx^{(j)}|y).
\end{equation}
\begin{equation}
\hbox{Denote}\;\;G(\bx^{(i)},\bx^{(j)}|y)= \frac{f(\bx^{(i)},\bx^{(j)}|y) f(\bx^{(i)}|y) f(\bx^{(j)}|y)}{f(\bx^{(i)},\bx^{(j)}|y)+\;f(\bx^{(i)}|y)f(\bx^{(j)}|y)}.
\end{equation}
Recall the conditional geometric mutual information (GMI) from \cite{SalimehEntropy2018}: for continuous random variables $\BX^{(i)}$, $\BX^{(j)}$ and discrete $Y$, the conditional GMI measure given $Y$ is defined by
\begin{equation}\label{Def:GMI}\begin{array}{ccl}
I(\BX^{(i)};\BX^{(j)}|Y)&=&\diy\mathbb{E}_{Y}\Big[I(\BX^{(i)};\BX^{(j)}|Y=y)\Big]\\[10pt]
&=&\diy\sum\limits_{y}p_y I(\BX^{(i)};\BX^{(j)}|Y=y),
\end{array}\end{equation}
where 
\begin{equation}\label{Def2:GMI}
I(\BX^{(i)};\BX^{(j)}|Y=y)=
\quad 1-2\diy\iint G(\bx^{(i)},\bx^{(j)}|y)\; d \bx^{(i)} d\bx^{(j)}.
\end{equation}
We define a measure based on joint probability densities $f_{ij}$ and $\pi_{ij}$:
\begin{equation}
    \delta_{yz}^{(ij)}=\diy\iint \frac{f(\bx^{(i)},\bx^{(j)}|y)f(\bx^{(i)}|z)f(\bx^{(j)}|z)}{f(\bx^{(i)},\bx^{(j)})+\pi(\bx^{(i)},\bx^{(j)})} \;d \bx^{(i)} d\bx^{(j)}.
\end{equation}
The following theorem derives a lower bound for $I(\BX^{(i)};\BX^{(j)}|Y)$. This bound will be used from now on instead of the conditional mutual information (CMI) to be maximized. For feature selection, we filter out the features with the highest CMI measures. 
\begin{theorem}\label{thm1}
Consider conditional probability densities $f(\bx^{(i)},\bx^{(j)}|y)$, $f(\bx^{(i)}|y)$, and $f(\bx^{(j)}|y)$ with priors $p_y$ $y=1,2,\ldots,m$. Then the conditional GMI between features $\BX^{(i)}$ and $\BX^{(j)}$ given label variable $Y$ is lower bounded by
\begin{equation}\label{lower.bound}
I(\BX^{(i)};\BX^{(j)}|Y)\geq 1-2\diy \sum\limits_{y}\sum\limits_{z} p_y p_z \delta_{yz}^{(ij)}.
\end{equation}
\end{theorem}
{\bf Proof:} We note that the conditional GMI $I(\BX^{(i)},\BX^{(j)}|Y=y)$ in (\ref{Def2:GMI}), this information can be written in terms of Henze-Penrose (HP) divergence \cite{HP} when $p=\diy\frac{1}{2}$ denoted by $D_{1/2}$:
\begin{equation}
I(\BX^{(i)},\BX^{(j)}|Y=y)=D_{1/2}\left(f(\bx^{(i)},\bx^{(j)}|y),f(\bx^{(i)}|y)f(\bx^{(j)}|y)\right).
\end{equation}
Therefore, the GMI in (\ref{Def:GMI}) is written in terms of $D_{1/2}$ as follows:
\begin{equation}\begin{array}{l}
I(\BX^{(i)},\BX^{(j)}|Y)=\diy\sum\limits_y p_y I(\BX^{(i)},\BX^{(j)}|Y=y)\\
=\diy\sum\limits_y p_y D_{1/2}\left(f(\bx^{(i)},\bx^{(j)}|y),f(\bx^{(i)}|y)f(\bx^{(j)}|y)\right). 
\end{array}\end{equation}
From \cite{TC} we know that the $f$-divergence function is  convex. Since HP-divergence belongs to the $f$-divergence family, we then have
\begin{equation}\label{eq1:proof1}\begin{array}{l}
I(\BX^{(i)},\BX^{(j)}|Y)\\
\quad \geq D_{1/2} \left(\diy\sum\limits_y p_y f(\bx^{(i)},\bx^{(j)}|y),\sum\limits_{y}p_yf(\bx^{(i)}|y)f(\bx^{(j)}|y)\right). 
\end{array}\end{equation}
By definition of joint probability densities $f_{ij}$ and $\pi_{ij}$ in (\ref{standard.joint}) and (\ref{markov.joint}) respectively, the RHS in (\ref{eq1:proof1}) is $D_{1/2} \left(f_{ij},\pi_{ij}\right)$. In addition we have 
\begin{equation}\begin{array}{l}
D_{1/2} \left(f_{ij},\pi_{ij}\right)
=1-2 \diy\sum_{y}\sum_{z} p_y p_z \delta^{(ij)}_{yz}.
\end{array}\end{equation}
This completes the proof of Theorem \ref{thm1}. \hfill$\square$

In the next section an estimation of the RHS of inequality (\ref{lower.bound}) is proposed.

\section{Estimation} \label{Estimate}
In this section, we introduce a new algorithm which estimates $\delta^{(ij)}_{yz}$ directly without requiring density estimates. The estimator is constructed based on the global MST \cite{SalimehetalAlertton2018}, and the random permuted sample \cite{SalimehEntropy2018}. The approach is called geometric conditional MI (GMI) and is summarized in Algorithm 1.
\begin{algorithm}[h]\label{algorithm1}
 \caption{Global FR estimator of $\delta^{(st)}_{yz}$}
 \begin{algorithmic}[1]
 \renewcommand{\algorithmicrequire}{\textbf{Input:}}
 \renewcommand{\algorithmicensure}{\textbf{Output:}}
 \REQUIRE Data set $\BZ_{2n}:=(\BX^{(s)}_{2n},\BX^{(t)}_{2n},\BY_{2n})$, $\big\{(\bx^{(s)}_i,\bx^{(t)}_i,\by_i)_{i=1}^{2n}\big\}$,
 $\BY\in\{y_1,y_2,\ldots,y_m\}$, $m=\# \{y_1,y_2,\ldots,y_m\}$\\
  \STATE Divide $\BZ_{2n}$ into two subsets $\BZ'_{n}$ and $\BZ''_{n}$\\
 \vspace{0.1cm}
     \STATE Partition $\BZ'_{n}$ into $m$ subsets $\BZ'_{n'_1|z_1},\BZ'_{n'_2|z_2},\ldots,\BZ'_{n'_m|z_m}$ according to $Y\in\{z_1,z_2,\ldots,z_m\}$ with sizes $n'_1,n'_2,\ldots,n'_m$
  \vspace{0.05cm}
  \STATE Partition $\BZ''_{n}$ into $m$ subsets based on different values $\{y_1,y_2,\ldots,y_m\}$ denoted by $\BZ''_{n|y_1},\BZ''_{n|y_2},\ldots,\BZ''_{n|y_m}$, such that $\BZ''_{n}=\bigcup\limits_{i=1}^m \BZ''_{n|y_i}$. Denote $n_i=\#\BZ''_{n|y_i}$ and $n=\diy\sum_{i}^m n_i$
 \vspace{0.05cm}
  \STATE $\widetilde{\BZ}_{n_i|y_i}\leftarrow \big\{(\bx^{(s)}_{y_i,k},\bx^{(t)}_{y_i,k})_{k=1}^{n_i}$ selected in random from $\BZ''_{n_i|y_i}\big\}$
  \vspace{0.05cm}
  \STATE $\widehat{\BZ}_{2n}\leftarrow \left(\bigcup\limits_{j}^m\BZ'_{n'_j|z_j}\right)\bigcup \left(\bigcup\limits_{i}^m\widetilde{\BZ}_{n_i|y_i}\right)=\BZ'_n\cup\widetilde{\BZ}_n$\\
  \vspace{0.05cm}
  \STATE Construct MST on $\widehat{\BZ}_{2n}$\\
  \vspace{0.05cm}
  \STATE $\mathfrak{R}_{z_j,y_i}\leftarrow\#$ edges connecting a node in $\BZ'_{n'_j|z_j}$ to a node of $\widetilde{\BZ}_{n_i|y_i}$\\
 \vspace{0.05cm}
  \STATE $\widehat{\delta}_{y_i,z_j}^{(st)}\leftarrow \left(\diy\frac{n}{2\; n'_j\;n_i}\right) \mathfrak{R}_{z_j,y_i}$
  \vspace{0.05cm}
  \ENSURE  $\widehat{\delta}_{y,z}^{(st)}$
 \end{algorithmic} 
 \end{algorithm}
 The next theorem asserts that our proposed estimator approximates $\widehat{\delta}_{y,z}^{(st)}$ when the number of samples from each label class tends to infinity. Due to space limitations only a sketch of the proof is given. Denote $\mathfrak{R}_{z_j,y_i}$ the FR statistic between original sample with label $z_j$ and random permutations sample with label $y_i$. 
 \begin{theorem}\label{thm:2}
Let $y\in\{y_1,\dots, y_m\}$ and $z\in\{z_1,\dots, z_m\}$, with $n'_j\rightarrow \infty$, $n_i\rightarrow \infty$, $n\rightarrow \infty$ such that $n'_j/n\rightarrow p_{z_j}$,  $n_i/n\rightarrow p_{y_i}$. Then
\begin{equation} \label{eq:V}
\left(\diy\frac{n}{2\; n'_j\;n_i}\right) \mathfrak{R}_{z_j,y_i} \diy\longrightarrow \delta^{(st)}_{y_i,z_j}\;\; \;\; \hbox{(a.s.)}
\end{equation}
\end{theorem}
{\bf Proof:} 
The following lemma will be required. It is a consequence of Lemma 5.2 in \cite{SalimehEntropy2018}. 
\begin{lemma}\label{lem1}
Consider the realization $\BZ_{2n}:=(\BX^{(s)}_{2n},\BX^{(t)}_{2n},\BY_{2n})$, $\big\{(\bx^{(s)}_i,\bx^{(t)}_i,\by_i)_{i=1}^{2n}\big\}$ where $Y$ is  a discrete random variable $Y$ taking values in the set $\{y_1,y_2,\ldots,y_m\}$. Following notations in Algorithm 1, let $\widetilde{\BZ}_{n_i|y_i}$ be a set having components selected at random from $\BZ''_{n_i|y_i}$. Then $n\rightarrow \infty$, for all classes $i\in\{1,\ldots,m\}$  the $\widetilde{\BZ}_{n_i|y_i}=\big\{(\bx^{(s)}_{y_i,k},\bx^{(t)}_{y_i,k})_{k=1}^{n_i}\big\}$ has conditional probability densities $f(\bx^{(s)}|y_i)f(\bx^{(t)}|y_i)$. 
\end{lemma}

Now to prove Theorem \ref{thm:2}, recall the sets $\BZ'_{n}$ and $\widetilde{\BZ}_{n}$ from Algorithm 1 and let $\BZ'_{n'_j|z_j}$ and $\widetilde{\BZ}_{n_i|y_i}$ have the joint probability densities $f(\bx^{(s)},\bx^{(t)}|z_j)$ and $\widetilde{f}(\bx^{(s)},\bx^{(t)}|y_i)$ such that from Lemma \ref{lem1} when $n$ tends to infinity and $n_i\rightarrow \infty$ then $\widetilde{f}(\bx^{(s)},\bx^{(t)}|y_i)\rightarrow f(\bx^{(s)}|y_i) f(\bx^{(t)}|y_i)$. 
Let $M_{n'_j}$ and $N_{n_i}$ be Poisson variables with mean $n'_j$ and $n_i$ respectively, for $i,j=1,2,\ldots, m$ and independent of one another and of $\BZ'_{n'_j|z_j}$ and $\widetilde{\BZ}_{n_i|y_i}$. Let $\widehat{\BZ}_{2n}=\BZ'_n\cup\widetilde{\BZ}_n$.
Denote $\mathfrak{R}_{z_j,y_i}$ the FR test statistic constructed by using global MST over $\BZ'_n\cup\widetilde{\BZ}_n$. 
Construct the global MST over $\left(\bigcup\limits_{j}^m\overline{\BZ}'_{z_j}\right)\bigcup \left(\bigcup\limits_{i}^m\overline{\BZ}_{y_i}\right)$. Let $\overline{\mathfrak{R}}_{z_j,y_i}:=\mathfrak{R}^{(ij)}(\overline{\BZ}'_{z_j},\overline{\BZ}_{y_i})$ be FR test statistic.  
It is sufficient to prove 
\begin{equation}\label{eq:thm3.2} \diy\frac{\;\mathbb{E}\big[\overline{\mathfrak{R}}_{z_j,y_i}\big]}{2n}\longrightarrow p_{z_j}p_{y_i}\delta^{(st)}_{z_j,y_i}.\end{equation}
\def\BW{\mathbf{W}}\def\bw{\mathbf{w}}
For $\bar{n}=(n'_1,n'_2,\ldots,n'_m, n_1,n_2,\ldots,n_m)$, such that $\sum\limits_{l}^m n'_l=\sum\limits_{l}^m n_l=n$, Let $\BW_1^{\bar{n}},\BW_2^{\bar{n}},\ldots$, be independent vectors with common densities, $ g_n(\bx^{(s)},\bx^{(t)})$
\begin{equation}\begin{array}{l}
    {(2n)}^{(-1)}\left(\sum\limits_{z_j} n'_jf(\bx^{(s)},\bx^{(t)}|z_j)+\sum\limits_{y_i} n_i\widetilde{f}(\bx^{(s)},\bx^{(t)}|y_i)\right). 
\end{array}\end{equation}
Next let $L_n$ be an independent Poisson variable with mean $2n$. 
We prove (\ref{eq:thm3.2}) for $\widetilde{\mathfrak{R}}_{z_j,y_i}$ which is the FR statistics for a $m$ marked points. 
\def\bz{\mathbf{z}}
Given points of $\mathfrak{F}'_{n}$ at $\bz=(\bx^{(s)},\bx^{(t)})$ and $\bar{\bz}=(\bar{\bx}^{(s)},\bar{\bx}^{(t)})$ the probability that they have different marks in $\{z_1,z_2,\ldots,z_m\}$ and $\{y_1,y_2,\ldots y_m\}$ is given by
\begin{equation} 
G_{n}(\bz,\bar{\bz}):=\diy\frac{n'_{j} f(\bz|z_j) n_{i} \widetilde{f}(\bar{\bz}|y_i)+n'_{j} f(\bar{\bz}|z_j) n_{i} \widetilde{f}(\bz|y_i)}
{K_n(\bz)\; K_n(\bar{\bz})},
\end{equation}
where 
$$K_n(\bz)=\sum\limits_{z_j} n'_j f(\bz|z_j)+\sum\limits_{y_i} n_i\widetilde{f}(\bz|y_i).$$
Set
\begin{equation}
    G(\bz,\bar{\bz}):=\diy\frac{p_{z_j} p_{y_i}\left(f(\bz|z_j)  \widetilde{\pi}(\bar{\bz}|y_i)+f(\bar{\bz}|z_j) \widetilde{\pi}(\bz|y_i)\right)} {K(\bz)\; K(\bar{\bz})},
\end{equation}
where 
$$K(\bz)=\sum\limits_{z_j} p_j f(\bz|z_j)+\sum\limits_{y_i} p_i\widetilde{\pi}(\bz|y_i),$$
and $\widetilde{\pi}(\bz|y_i)=f(\bx^{(s)}|y_i)f(\bx^{(t)}|y_i)$.
We observe that $G_{n}(\bz,\bar{\bz})\rightarrow G(\bz,\bar{\bz})$ as they range in $[0,1]$. For $z_j<y_i$
\begin{equation}\label{eq2:1.1}\begin{array}{l}
\mathbb{E}\left[\widetilde{\mathfrak{R}}_{z_j,y_i}|\mathfrak{F}'_n\right]=\diy\mathop{\sum\sum}_{\ 1\leq t<l\leq L_{n}} G_{n}(\BW^{\bar{n}}_t,\BW^{\bar{n}}_l)
\mathbf{1}\big\{(\BW^{\bar{n}}_t,\BW^{\bar{n}}_l)\in \mathfrak{F}'_n\big\}.
\end{array}\end{equation}
By taking the expectation (\ref{eq2:1.1}) we can write
\begin{equation}\begin{array}{l}
\diy\mathbb{E}\left[\widetilde{\mathfrak{R}}_{z_j,y_i}\right]=o(2n)\\[10pt]
+\diy\mathbb{E}\mathop{\sum\sum}_{\ 1\leq t<l\leq L_{n}} G(\BW^{\bar{n}}_t,\BW^{\bar{n}}_l)
\mathbf{1}\big\{(\BW^{\bar{n}}_t,\BW^{\bar{n}}_l)\in \mathfrak{F}'_n\big\}
.\end{array}\end{equation} 
By taking into account the non-Poisson process, 
we have 
\begin{equation} \begin{array}{l}
\diy \mathbf{E}\left[\widetilde{\mathfrak{R}}_{z_j,y_i}\right]=o(2n)\\[10pt]
+\diy\mathbb{E}\mathop{\sum\sum}_{\ 1\leq t<l\leq 2n} G(\BW^{\bar{n}}_t,\BW^{\bar{n}}_l) \mathbf{1}\big\{(\BW^{\bar{n}}_t,\BW^{\bar{n}}_l)\in \mathfrak{F}_n\big\}
.\end{array}\end{equation}
Introduce $g(\bx^{(s)},\bx^{(t)})$ as 
    \begin{equation}\begin{array}{l}
    ={(2)}^{(-1)}\left(\sum\limits_{z_j} p_jf(\bx^{(s)},\bx^{(t)}|z_j)+\sum\limits_{y_i} p_i f(\bx^{(s)}|y_i) f(\bx^{(t)}|y_i)\right). 
\end{array}\end{equation}
We can write that $g_n(\bx^{(s)},\bx^{(t)})\rightarrow g(\bx^{(s)},\bx^{(t)})$. Consequently by proposition 1 in \cite{HP}, we have
\begin{equation*} \begin{array}{l}
\diy\mathbb{E}\left[\widetilde{\mathfrak{R}}_{z_j,y_i}\right]\big/2n\rightarrow\int G(\bz,\bz) g(\bz)\rd\bz
\end{array}\end{equation*}
And this proves Theorem \ref{thm:2}. \hfill$\square$ 
\vspace{-0.6cm}
\section{Experiments}\label{Experiments}
\begin{figure}[t]\label{Fig.1}
 \vspace{-0.6cm}
 \centering
\includegraphics[width=2.8in]{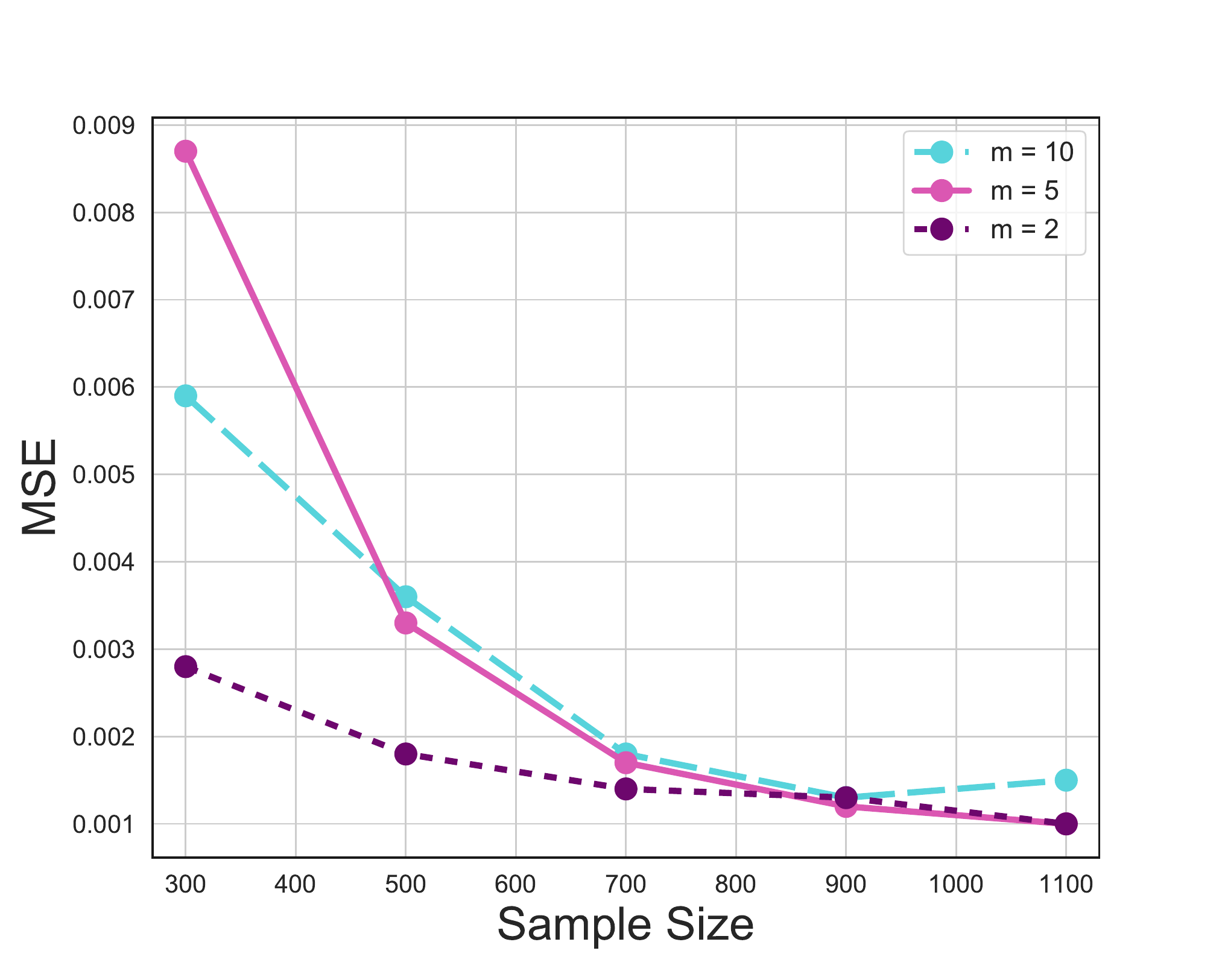}
\vspace{-0.2cm}
\caption{Convergence in  MSE  of CMI lower bound estimator for $d=2$. The MSE for samples drawn from three sets of distributions with $m=2,5,10$ normal distributions in each set ($\mu=0.5$). The results were averaged over 50 iterations.}
\vspace{-0.2cm}
\label{fig1}
\end{figure}
In this section we perform multiple experiments to validate our theory and to demonstrate the utility of our proposed feature selection algorithm on MNIST  data set. We use the maximum lower bound on the conditional mutual information in (\ref{Def:GMI}) as a proposed measure to filter irrelevant features. In the following simulation we first analyze the proposed estimator of the CMI lower bound. We draw samples from three sets of distributions with $m=2,5,10$ normal distributions $\mathcal{N}(\mu_i, 0.1I)$ in each set, where $\mu=0.5$ and $\mu_i$ as in Section IV \cite{SalimehetalAlertton2018}.
The sample size for all classes are equal. Fig~\ref{fig1} shows  the  MSE  between  the  estimated  and  oracle lower bound on CMI as a function of total sample size $N$, for different labels $m=2,5,10$.  The behavior of MSE in terms of $N$ is clear but note that as the number of labels grow the MSE increases, too. This means that as the number of labels increases, more samples are required to estimate the CMI bound accurately. 

One of our main motivations for the GMI for feature selection is less computational complexity. Hence an experiment was performed, Fig~\ref{fig2}, to compare the runtime of our method with that of Berisha et al's method \cite{Berishaetal2016}.
\begin{figure}[t]\label{Fig.3}
 \vspace{-0.3cm}
 \centering
\includegraphics[width=2.8in]{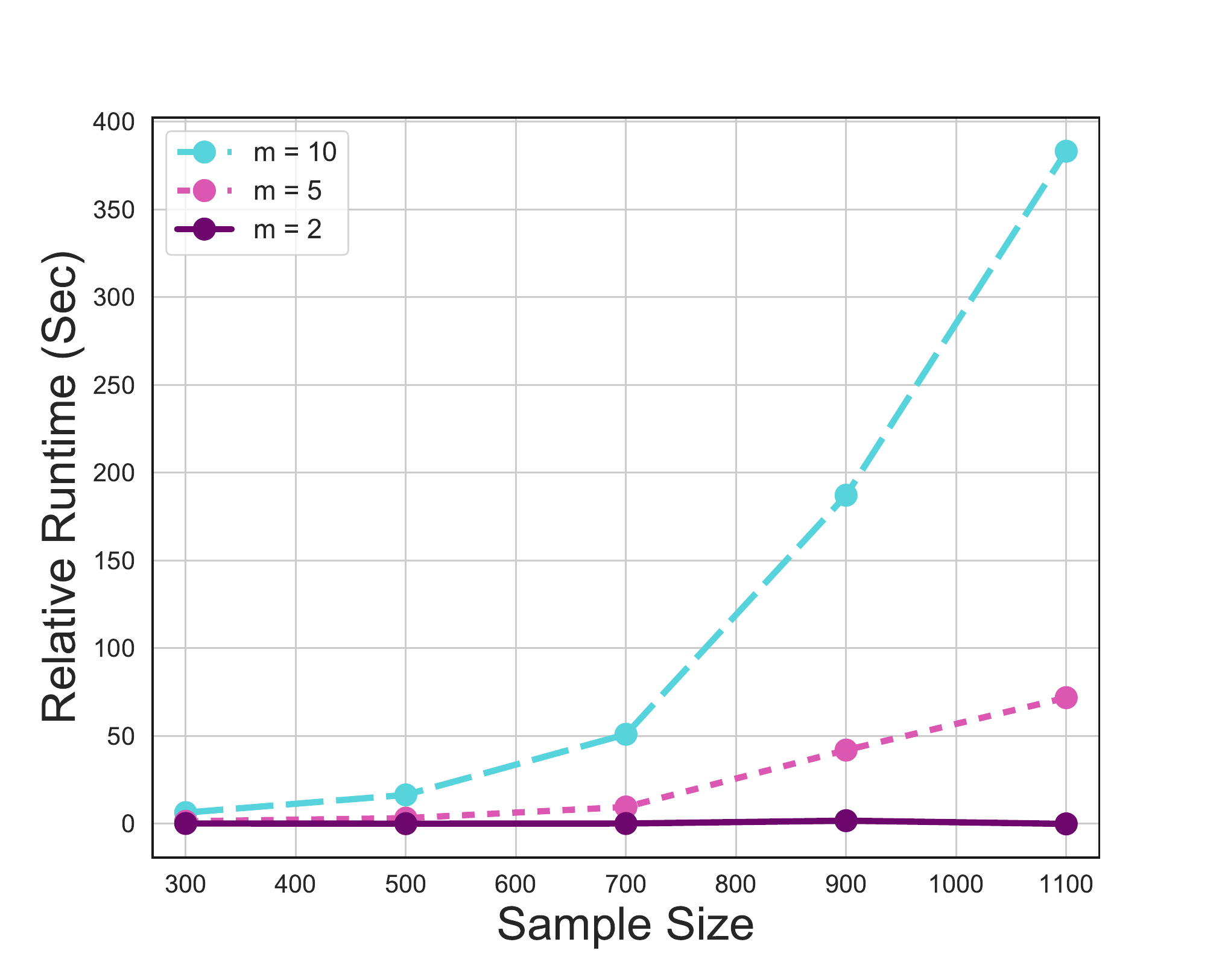}
\vspace{-0.2cm}
\caption{Relative runtime of pairwise Dp method for m classes and proposed Geometric CMI algorithm vs. sample size. For large class our proposed method offers significantly faster runtime.}
\vspace{-0.5cm}
\label{fig2}
\end{figure}
Subsequently, we utilize our proposed method to explore feature selection for the MNIST data set. The MNIST data set consists of grey-scale images, 28 x 28 pixels, of hand-written digits 0 - 9.
In this experiment, first we use PCA on 2000 training images to reduce the feature  dimensions and detect latent features.  We consider the first 30 principle components and implement our proposed method on this subset. The pairwise Dp algorithm where the training and test data have the same distribution is applied to select features with minimum total multi-class Bayes error upper bound given in Theorem 2 \cite{Berishaetal2016}. 
Next we show the results of applying the CMI estimator and two state-of-the-art feature selection methods Linear Support Vector Classification (LSVC)~\cite{Branketal2002, Bietal2003} and tree-based method (Extra-Tree-Classifier (ETC))~\cite{AuretAldrich2011} to MNIST data sets for various sample sizes $N=100,300,400$. Table 1 shows the general FR test statistics (denoted by GMI) and total estimated pairwise upper bounds for Bayes Error (denoted by Dp). Fig \ref{fig-3} shows a comparison of the average classification as a function of feature set size between GMI approach, Dp criteria, LSVC, and ETC approaches. We applied a multi-class SVM to measure classification accuracy on training data of size $10^4$. The generalized FR test statistic for smaller size of feature sets, selects the features with higher average accuracy (after 10 runs). Observe that both GMI and Dp approaches outperform LSVC and ETC methods.
\vspace{-2pt}
\begin{table}[h]
{\small
\begin{center}
\scalebox{0.8}{
\begin{tabular}{ |c||c|c|  }
\hline
Number of Features & Algorithms  & Number of Training Sample \\
& & {100}\;\; \;\; \; \;\; 300\;\; \; \;\; {500}\\
 \hline
10 & GMI  &  {{\bf 61.48}}\;\; \; {\bf 61.47}   \;\; \; {\bf 60.43} \\
 & Dp  & {57.31}\;\; \; 51.57      \;\; \; 55.53  \\
 & LSVC  & {20.00}\;\; \;  5.99       \;\; \; \;\; {8.40}  \\
  & ETC  & {10.69 }\;\; \;  6.00        \;\; \; \;\; {7.09}  \\
15 & GMI  &  {70.01}\;\; \; 69.94   \;\; \; 66.48\\
 & Dp  & {64.86}\;\; \;69.90    \;\; \; 71.71   \\
 & LSVC  & {22.26}\;\; \;  9.86       \;\; \; \;\;{10.51}  \\
  & ETC  & {22.26}\;\; \;  9.84        \;\; \; \;\; {10.51}  \\
20 & GMI &  {73.99}\;\; \;  73.94  \;\; \; 72.27\\
 & Dp  &  {78.95}\;\; \; 77.83   \;\; \; 76.77\\
 & LSVC  & {22.4}\;\; \;  9.92       \;\; \; \;\;{13.42}  \\
   & ETC  & {24.67}\;\; \; 9.93        \;\; \; \;\; {12.77}  \\
 \hline
\end{tabular}
}
\caption {{Average Classification Accuracies of Top Features Selected by GMI and pairwise Dp statistic.}}
\end{center}
\label{tab:real_data}}
\end{table}

\begin{figure}[!h]
 \vspace{-1.1cm}
 \centering
\includegraphics[width=2.7in]{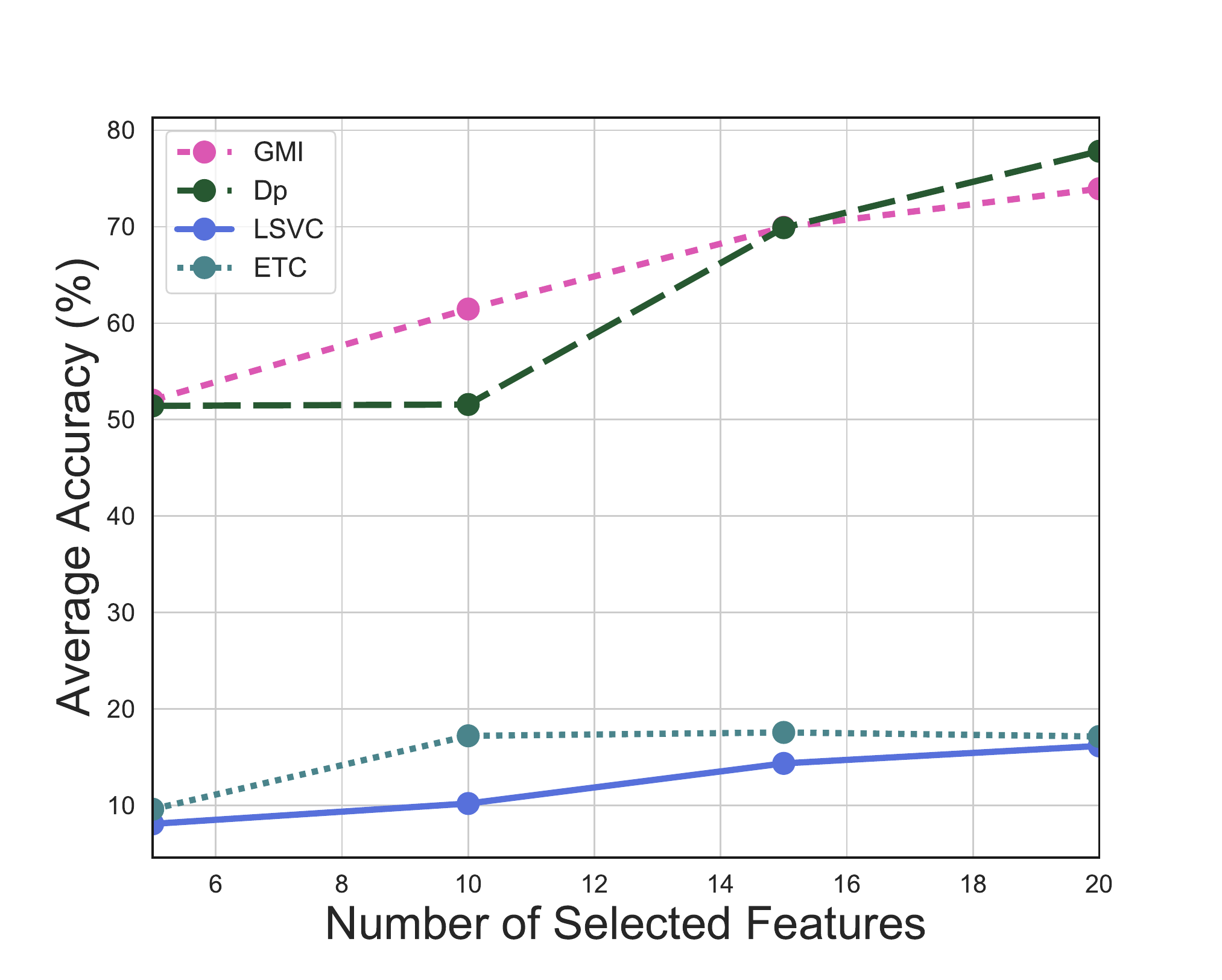}
 \vspace{-0.4cm}
\caption{A comparison of the average classification as a function of feature set size using both the GMI and Dp criteria.}
 \vspace{-0.5cm}
\label{fig-3}
\end{figure}
\vspace{-0.1cm}
\section{Conclusion}\label{conclude}

In this paper, we proposed a new technique to select features via maximizing conditional dependency between features given the variable. We derived a lower bound for a geometric dependency measure, the conditional geometric mutual information (GMI). We showed that using a global FR statistic derived from a global MST the lower bound can be estimated directly. This estimator has low computational complexity.
We demonstrated that our proposed algorithm and Dp estimator are more accurate than LSVC and ETC methods when applied to feature selection on the MNIST data set. 
\newpage
\label{sec:refs}
\bibliographystyle{IEEEbib}
\bibliography{refs}

\end{document}